\documentclass[conference]{IEEEtran}
\IEEEoverridecommandlockouts

\usepackage{cite}
\usepackage{amsmath,amssymb,amsfonts}
\usepackage{algorithmic}
\usepackage{graphicx}
\usepackage{textcomp}
\usepackage{xcolor}
\usepackage{float}
\usepackage[utf8]{inputenc}
\usepackage[T5]{fontenc}
\usepackage{multirow}

\def\BibTeX{{\rm B\kern-.05em{\sc i\kern-.025em b}\kern-.08em
    T\kern-.1667em\lower.7ex\hbox{E}\kern-.125emX}}
\begin{document}

\title{Predicting Job Titles from Job Descriptions with Multi-label Text Classification\\
}

\author{\IEEEauthorblockN{Hieu Trung Tran}
\IEEEauthorblockA{\textit{University of Information Technology} \\
\textit{Vietnam National University}\\
Ho Chi Minh City, Vietnam \\
18520754@gm.uit.edu.vn}
\and
\IEEEauthorblockN{Hanh Hong Phuc Vo}
\IEEEauthorblockA{\textit{University of Information Technology} \\
\textit{Vietnam National University}\\
Ho Chi Minh City, Vietnam  \\
18520275@gm.uit.edu.vn}
\and
\IEEEauthorblockN{Son T. Luu}
\IEEEauthorblockA{\textit{University of Information Technology} \\
\textit{Vietnam National University}\\
Ho Chi Minh City, Vietnam \\
sonlt@uit.edu.vn}
}

\maketitle

\begin{abstract}

Finding a suitable job and hunting for eligible candidates are important to job seeking and human resource agencies. With the vast information about job descriptions, employees and employers need assistance to automatically detect job titles based on job description texts. In this paper, we propose the multi-label classification approach for predicting relevant job titles from job description texts, and implement the Bi-GRU-LSTM-CNN with different pre-trained language models to apply for the job titles prediction problem. The BERT with multilingual pre-trained model obtains the highest result by F1-scores on both development and test sets, which are 62.20\% on the development set, and 47.44\% on the test set.
\end{abstract}

\begin{IEEEkeywords}
job descriptions, job titles, multi-label classification, deep neural models, transformer models.
\end{IEEEkeywords}

\section{Introduction}
\label{intro}
Recently, the development of online job seeking make opportunities for many employees access to the useful information for their jobs. Thereby, recruiters find suitable candidates by posting job description and requirements for each job title. However, with the vast information of job descriptions and requirements, it is necessary to categorize this information into specific job titles. This is not only used for employees to easily find the most suitable job, but also can recommend relevant job titles for employees. Therefore, we propose a method to automatically detect job titles based on job descriptions posts.

The task of predicting suitable job titles from job descriptions is a text classification task. This task takes the input as a post of job descriptions, then predicts the suitable job titles for job description post as the output. However, the output for a job description post are not only one job title but also several of related job titles corresponding to the job description content. Hence, we apply the multi-label classification approach for the task of predicting job titles from job description.

To implement this task, we firstly collect the data about job descriptions and job requirements from various top online job finding websites in Vietnam. After collected the dataset, we implement the deep learning models for multi-label text classification on the dataset to make predict job titles from job description. Based previous works about Job prediction \cite{9140760} and Bi-GRU-LSTM-CNN model \cite{van2019hate}, we applied to our task for predicting suitable job titles according to job description texts. Our main contribution in this paper is providing a dataset about job titles prediction and proposing a method to predict suitable job titles based on job description by multi-label text classification approach.

The paper is structured as follows. Section \ref{related_works} investigates available methodologies and techniques for the job prediction task and multi-label text classification approach. Section \ref{dataset} overviews about our dataset construction and characteristics of the dataset. Section \ref{methodologies} proposes our methodologies for job titles prediction based on multi-label text classification approach. Section \ref{experiemnts} shows our empirical results on the dataset. Finally, Section \ref{conclusion} concludes our works.

\section{Related works}
\label{related_works}
Huynh et al. \cite{9140760} conducted an empirical study about job prediction: from deep neural network models to applications. In this paper, the authors focus on studying the job prediction using different deep neural network models including TextCNN, Bi-GRU-LSTM-CNN, and Bi-GRU-CNN. Their experimental results illustrated that our proposed ensemble model achieved the highest result with an F1-score of 72.71\%. Also, Huynh et al. \cite{van2019hate} proposed the Bi-GRU-LSTM-CNN model for the task of hate speech detection and achieved optimistic result. However, these two tasks are multi-class classification, in which the predictions by the classification model are just single labels. Hence, we propose our methodologies based on those previous works for our task.

Bhola et al. \cite{bhola-etal-2020-retrieving} proposed a multi-label classification method for the task of retrieving relevant skills based on job requirements, which is similar to our problem. From the obtained results, BERT \cite{devlin-etal-2019-bert} gives the highest results, which indicates the power of BERT and its variances for this task. Therefore, BERT and other variances including XLM-R \cite{conneau-etal-2020-unsupervised}, DistilBERT \cite{sanh2019distilbert} and PhoBERT \cite{2020-phobert} - a Vietnamese pre-trained model based on RoBERTa \cite{liu2019roberta} are also our chosen approaches for the task of job titles prediction.

F1-score are used in both multi-class classification and multi-label classification because of its ability to treat all classes in the dataset and avoid the bias among classes when evaluating the results \cite{sorower2010literature,sokolova2009systematic}. Hence, we use the F1 score for evaluating the performance of classification models for the task of job titles predictions.

\section{Dataset}
\label{dataset}
\subsection{Data collection}

\begin{table*}[ht]
    \centering
    \caption{Example of job descriptions and job titles from the dataset}
    \label{tbl_example_jobs}
    \begin{tabular}{|c|p{13.5cm}|p{3cm}|}
        \hline
        \textbf{\#} & \textbf{Job description} & \textbf{Job} \\
        \hline
        1 & Cung cấp trợ giúp cho khách hàng sử dụng các sản phẩm hoặc dịch vụ Giao tiếp lịch sự với khách hàng qua điện thoại, email, thư từ và các kênh online khác Xem xét và giải quyết các vấn đề của khách hàng, có thể phức tạp hoặc lâu dài, đã được các trợ lý dịch vụ khách hàng chuyển giao Xử lý khiếu nại của khách hàng hoặc bất kỳ sự cố lớn nào, chẳng hạn như vấn đề bảo mật hoặc khách hàng bị ốm Phân tích số liệu thống kê hoặc dữ liệu khác để xác định mức độ dịch vụ khách hàng Tham mưu Cải thiện các thủ tục, chính sách và tiêu chuẩn dịch vụ khách hàng Phối hợp tốt với các bộ phận khác để thảo luận về những cải tiến có thể có đối với dịch vụ khách hàng (\textbf{English:} Provide help to customers using products or services Communicate politely with customers by phone, email, mail, and other online channels Review and resolve customer issues, which can be complex complex or lengthy, provided by a customer service assistant Handle customer complaints or any major issues, such as security issues or sick customers Analyze statistics or data other data to determine customer service levels Advise on improvements to customer service standards, policies and procedures Work well with other departments to discuss possible improvements to with customer service) & Tư vấn (Consulting), Dịch vụ khách hàng (Customer Service), Bán hàng / Kinh doanh (Sales/Business) \\
        \hline
        2 & Tham gia vào các dự án triển khai ERP Oracle (Cloud \& On-Premise) Thực hiện khảo sát nghiệp vụ, tiếp nhận và phân tích các yêu cầu từ khách hàng. Tham gia thực hiện việc kỹ thuật, thiết kế báo cáo từ giải pháp ERP Oracle (Cloud \& On-Premise). Hướng dẫn/ đào tạo và chuyển giao kỹ thuật cho khách hàng sử dụng hệ thống ERP Oracle (Cloud \& On-Premise). Hỗ trợ vận hành sau khi triển khai. (\textbf{English:} Participate in ERPOracle implementation projects (Cloud \& On-Premise) Conduct business surveys, receive and analyze requests from customers. Participate in technical implementation, report design from Oracle ERP solution (Cloud \& On-Premise).Instruction/training and technical transfer for customers to use ERPOracle system (Cloud \& On-Premise). Operational support after deployment.) 
        & CNTT - Phần mềm (IT - Software) \\
        \hline
        3 & - Phối hợp với tổ mua hàng kiểm tra hóa đơn và các chứng từ đầy đủ kèm theo hàng hóa; - Lập báo giá, hợp đồng, đơn đặt hàng, hồ sơ bán hàng; - Thực hiện việc đối chiếu vật tư hàng hóa giữa các công trình; - Theo dõi, xử lý công nợ phải thu, thực hiện tính lãi quá hạn; - Thực hiện các công việc khác theo yêu cầu liên quan đến công tác bán hàng, công nợ khách hàng; (\textbf{English:} Coordinating with the purchasing team to check invoices and complete documents attached to the goods; - Making quotations, contracts, purchase orders, sales records; - Perform the comparison of materials and goods between the works; - Monitor and handle receivables, calculate overdue interest; - Perform other tasks as required related to sales, customer debt) 
        & Thống kê, Bán hàng (Statistics, Sales) / Kinh doanh (Business), Kế toán / Kiểm toán (Accounting / Auditing / Tax)\\
        \hline
    \end{tabular}
\end{table*}

We collect the textual data about job seeking on several of online Vietnamese job seeking sites. The data includes job descriptions, job requirements and job titles. However, the job titles are different on each site. Therefore, we have to normalize the job titles manually after collected data from different sites to a standard list of jobs. This list contains totally 68 job titles (See Appendix \ref{appendix_job_title} for details). Table \ref{tbl_example_jobs} illustrates examples from our dataset.

\subsection{Data preparation}
We collect approximately 22,000 job descriptions from different site to construct the training set. However, to guarantee the objective of the dataset, we collect nearly 4,000 other job descriptions which are different from previous data in the training set to construct the test set. Then, we take randomly 10\% from each to create the development set. Table \ref{tbl_overview_detail} shows details information about each set. 

\begin{table}[H]
    \centering
    \caption{Detailed information about the dataset}
    \label{tbl_overview_detail}
    \begin{tabular}{|c|c|}
        \hline
        \textbf{} & \textbf{Number of job description} \\
        \hline
        Training set & 20,234 \\
        \hline
        Development set & 1,760 \\
        \hline
        Test set & 3,933 \\
        \hline
    \end{tabular}
\end{table}

\subsection{Data overview}
\begin{table}[H]
    \centering
    \caption{The length of job description in the dataset}
    \label{tbl_job_desc_detail}
    \begin{tabular}{|c|r|r|}
        \hline
        \textbf{} & \textbf{Maximum length} & \textbf{Average length}\\
        \hline
        Training set & 1,589 & 158.27 \\
        \hline
        Development set & 1,061 & 153.78 \\
        \hline
        Test set & 899 & 143.98\\
        \hline
    \end{tabular}
\end{table}

\begin{figure}[H]
    \centering
    \includegraphics[scale=0.55]{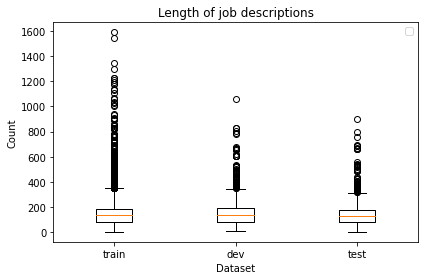}
    \caption{Distribution of text length in the dataset.}
    \label{fig_text_len_distr}
\end{figure}

Table \ref{tbl_job_desc_detail} shows the length of job description texts in both training, development, and test set in the dataset. The lengths are calculated by character levels. It can be seen that, although the maximum length in the test set and the development set is lower than the training set, the average text length of three parts are nearly the same. Figure \ref{fig_text_len_distr} illustrates the distribution of text length in training, development, and test set respectively.

Besides, as mentioned in Section \ref{intro}, the task of job title prediction is the multi-label text classification task, since each job description can have more than one label. Table \ref{tbl_label_details} summarizes the labels distribution for each job description in training, development, and test sets. It can be seen that, the proportion of job description that contains single label and job descriptions with more than one labels is nearly same, and number of job description with more than one label in the test set and development set are larger than in the training set. This is the challenge for classification models for giving corrects relevant job titles based on job descriptions.

\begin{table}[H]
    \centering
    \caption{Labels appeared in the dataset for each Job description (JD)}
    \label{tbl_label_details}
    \begin{tabular}{|c|c|c|c|}
        \hline
        \textbf{} & \textbf{Train} & \textbf{Dev} & \textbf{Test} \\
        \hline
        JD has 1 label & 10,380 & 734 & 986 \\
        \hline
        JD has more than 1 label & 9,854 & 1,026 & 2,947 \\
        \hline
        Maximum labels per JD & 7 & 5 & 7 \\
        \hline
    \end{tabular}
\end{table}

\begin{figure}[H]
    \centering
    \includegraphics[scale=0.4]{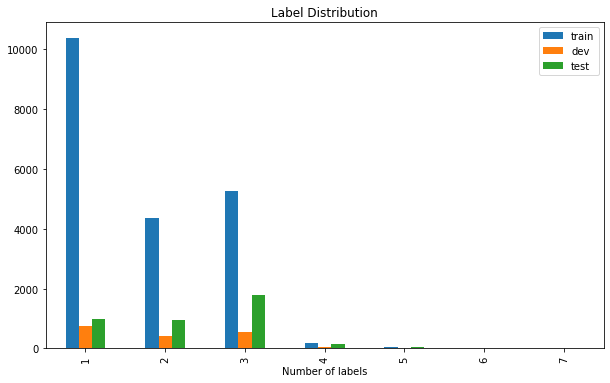}
    \caption{Number of labels per job descriptions in the dataset.}
    \label{fig_label_distr}
\end{figure}

In addition, according to Table \ref{tbl_label_details}, the maximum labels per job description in the training and test sets are 7, while the development set has the maximum of 5 labels per job description. To make it clear, Figure \ref{fig_label_distr} illustrates the amount of labels belongs to each job description text in the dataset. According to Figure \ref{fig_label_distr}, the majority of job descriptions contains one, two, and three labels, respectively.

Finally, in the dataset, the job descriptions are written in both Vietnamese and English. In the training set, there are 17,324 job description in Vietnamese, and 2,910 in English. In the development set, there are 1,508 Vietnamese job description, and 252 English ones. Last, the test set contains 3381 job descriptions written in Vietnamese, and 552 job descriptions written in English.

\section{Methodologies}
\label{methodologies}

\subsection{The model for multi-label classification}

The problem of our task is defined as below, which is categorized as multi-label classification task:
\begin{itemize}
    \item \textbf{Input:} Job descriptions as texts.
    \item \textbf{Output:} List of job titles that suitable to the input job descriptions.
\end{itemize}

\begin{figure}[H]
    \centering
    \includegraphics[scale=0.45]{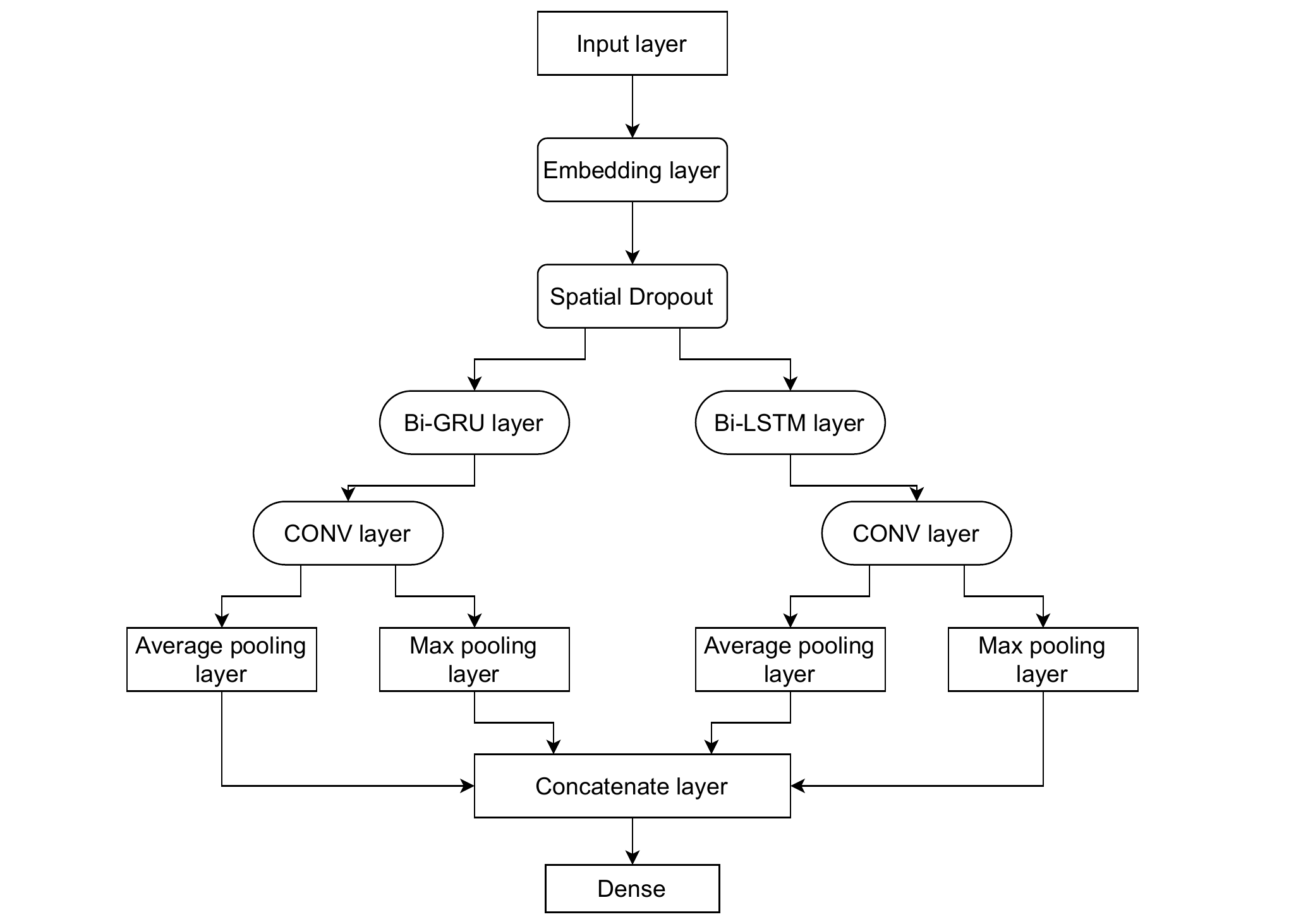}
    \caption{The architecture of the Bi-LSTM-GRU-CNN model.}
    \label{fig_classification_model}
\end{figure}

The Bi-GRU-LSTM-CNN model was used in the task of Job Prediction \cite{9140760} and Hate speech detection \cite{van2019hate}, and obtained optimistic results. The Bi-GRU-LSTM-CNN is described in Figure \ref{fig_classification_model}. Nevertheless, our problem is categorized as the multi-label classification task, which is different with the multi-class classification task in \cite{9140760}. Instead of returning the job title with the highest probability among list of job titles, the output of our task is the list of probabilities corresponding to the appearance of each job titles. Therefore, the labels of our problem are represented as one-hot vectors with 68 dimension, corresponding to 68 job titles in our dataset. We use the sigmoid function in the final layer (Dense layer) to control the probabilities of each label. If the probability of a label is larger than 0.5, then the label is set to 1, which is the chosen label. 

\subsection{Pre-trained language models}
In many of NLP tasks, word vector representation plays an important role, which can enhancing the performance of classification models by boosting the ability of capturing and understanding words in contexts \cite{10.1145/3373017.3373028,pham2017end}. Among all, pre-trained word embedding achieved state-of-the-art results in many of down stream task in NLP \cite{10.1145/3373017.3373028}, especially the appearance of BERT \cite{devlin-etal-2019-bert} and it variances. Hence, in this paper, we use the pre-trained models in Vietnamese including W2V, fastText, Bert\_Base, ELMO and MULTI\_WC\_F\_E\_B provided by Vu et al. \cite{vu-xuan-etal-2019-etnlp}, Pho2WV \cite{tuan-nguyen-etal-2020-pilot} as well as state-of-the-art pre-trained transformers model for Vietnamese languages such as multilingual BERT (m-BERT) \cite{devlin-etal-2019-bert}, XLM-R \cite{conneau-etal-2020-unsupervised}, DistilBERT \cite{sanh2019distilbert}, and PhoBERT \cite{2020-phobert} to evaluate the performance of job title prediction task in our dataset. In Figure \ref{fig_classification_model}, the embedding layer is built based on those mentioned pre-trained models. 

In addition, according to Thin et al. \cite{van2021investigating}, the word-level models better than syllable-level models on Vietnamese text classification. Hence, we use the VnCoreNLP \cite{vu-etal-2018-vncorenlp} to segment Vietnamese texts into words before fitting to the classification models.

\subsection{Evaluation metrics}
Assuming that $y$ is a one-hot vector lists with 68 elements corresponding to 68 job titles denoted for the true labels, and $\hat{y}$ denoted for prediction labels given by the classification models. With N prediction, The F1-score (according to \cite{sorower2010literature}) is calculated as: 
\begin{equation}
    \centering
    F{1} = \frac{1}{N}\sum_{i=1}^{N} \frac{2|y_{i} \cap \hat{y}_{i}|}{|y_{i}| + |\hat{y}_{i}|}
\end{equation}

\section{Experiments}
\label{experiemnts}
\subsection{Experimental settings and Data pre-processing}
We set the hyper-parameters of the Bi-GRU-LSTM-CNN model as: maximum length (max\_len) equals to 200, batch sizes equals to 256 (we use batch\_size equals to 32 for transformers pre-trained models), 100 units for both GRU, LSTM layers, and 50 units for 2 convolutional layers. Besides, we use the Adam optimizer and the Binary Crossentropy loss function. 

For the pre-trained word embeddings, we use those which are provided by Vu et al. \cite{vu-xuan-etal-2019-etnlp} at GitHub page\footnote{https://github.com/vietnlp/etnlp} and PhoW2V\footnote{https://github.com/datquocnguyen/PhoW2V} provided by Nguyen et al. at \cite{tuan-nguyen-etal-2020-pilot}. The dimension of word embeddings are 100 for PhoW2V, 300 for W2V and FastText, 768 for Bert\_Base, 1024 for ELMO, and 2392 for MULTI\_WC\_F\_E\_B. 

For the transformers pre-trained models, we use \textit{bert-multilingual-cased} for m-BERT, xlm-r-based for XLM-R, distilbert-base-cased for DistilBERT, and phobert-base for PhoBERT. All of these model are provided at Huggingface\footnote{https://huggingface.co/}.  

Finally, on the data pre-proceesing step, we remove special characters from the job descriptions texts, and segment texts into words by the VnCoreNLP toolkit\footnote{https://github.com/vncorenlp/VnCoreNLP}.

\subsection{Empirical results}

\begin{table}[H]
    \centering
    \caption{Results of the Bi-LSTM-GRU-CNN models on different of pre-trained language models}
    \label{tab_empirical_results}
    \begin{tabular}{|p{3cm}|l|r|r|}
        \hline
        \multirow{2}{*}{\textbf{Type}} & \multirow{2}{*}{\textbf{Pre-trained models}} & \multicolumn{2}{|c|}{\textbf{F1-score (\%)}} \\
        \cline{3-4}
        & & Dev & Test \\ 
        \hline
        \multirow{6}{*}{Pre-trained embedidings} & PhoW2V & 54.91 & 37.91 \\
        \cline{2-4}
        & W2V & 53.27 & 34.61 \\
        \cline{2-4}
        & fastText & 53.48 & 35.99 \\
        \cline{2-4}
        & Bert\_base & 56.96 & 41.31 \\
        \cline{2-4}
        & ELMO & 57.65 & 41.89 \\
        \cline{2-4}
        & \textbf{MULTI\_WC\_F\_E\_B} & \textbf{58.58} & \textbf{43.91} \\
        \hline
        \multirow{4}{*}{Transformer models} & \textbf{BERT} & \textbf{62.20} & \textbf{47.44} \\
        \cline{2-4}
        & XLM-R & 61.39 & 46.97 \\
        \cline{2-4}
        & DistilBERT & 57.85 & 42.16 \\
        \cline{2-4}
        & PhoBERT & 61.29 & 46.85 \\ 
        \hline
    \end{tabular}
\end{table}

Table \ref{tab_empirical_results} illustrates empirical results by the Bi-LSTM-GRU-CNN models on different pre-trained models. From the results, it can be seen that Transformer models obtained higher performance than pre-trained embeddings. The MULTI\_WC\_F\_E\_B showed highest results on the dataset among other word embeddings. For the transformer models, two multilingual models includes BERT and XLM-R showed the highest results, better than PhoBERT, which is a monolingual models trained on Vietnamese texts. This is because our dataset contains both job description written in Vietnamese and English as described in Section \ref{dataset}. However, the difference between BERT, XLM-R and PhoBERT is not too much.

\subsection{Error analysis}
Table \ref{tbl_error_job_title} describes the number of correct and wrong job titles prediction by the classification model. It can be seen that, for job description with single label, the correct predictions is better than job description with multiple labels. The more labels the data has, the less correct the models predict. This is the challenge for the classification model in the task of multi-labels classification, because the model not only give single correct prediction, but also giving full correct predictions relevant to each job description. Specially, the currents model cannot give correct prediction for job descriptions which labels contains more than 5 job titles.

\begin{table}[H]
    \centering
    \caption{Number of correct and wrong job titles predictions from the results by classification models}
    \label{tbl_error_job_title}
    \begin{tabular}{|c|c|c|}
        \hline
        \textbf{Number of job titles} & \textbf{\# Corrects} & \textbf{\# Wrongs} \\
        \hline
        1 & 560 & 426 \\
        \hline
        2 & 249 & 710 \\
        \hline
        3 & 89 & 1684 \\
        \hline
        4 & 19 & 131 \\
        \hline
        5 & 1 & 59 \\
        \hline
    \end{tabular}
\end{table}

In addition, Table \ref{tbl_incorrect_samples} describes several incorrect prediction sample from the results. In the first job description from Table \ref{tbl_incorrect_samples}, the classification model cannot recognize enough information from the texts, e.g. \textit{Manulife} - the name of the insurance company, to identify the job title about \textit{Insurances}. In addition, also from the job description No \#1, we can see that it describes the job as \textit{"...collecting, instructing and preparing customers' information and documents ..."}, which is a job about Customer services. However, the classification model cannot predict \textit{Customer services} for the given job description. Besides, the job description No \#2 confuses the classification models when the predicted job title is not relevant to any of the real job titles of the job description. This is because the entities \textit{"sales"}, \textit{"business"}, \textit{"debt"} are mostly mentioned in the content of the job description. The job description mentioned the \textit{"mechanism machines"} in the first sentence (see Job description No \#2 from Table \ref{tbl_incorrect_samples}), which is relevant to the job title named as \textit{Mechanical / Auto / Automotive}, and \textit{"chemical"}, which is relevant to \textit{Oil / Gas} or \textit{Chemical Eng}. Particularly, for the third job description (in Table \ref{tbl_incorrect_samples}), the classification model cannot predict relevant job titles. Hence, to boost up the ability of classification model, we need to integrate extra information about the job such as job names and job requirements with the job description. 


\begin{table*}
    \centering
    \caption{Several incorrect samples from the results}
    \label{tbl_incorrect_samples}
    \begin{tabular}{|c|p{2cm}|p{9cm}|p{2.3cm}|p{2.2cm}|}
        \hline
        \textbf{No.} & \textbf{Job names} & \textbf{Job descriptions} & \textbf{True labels} & \textbf{Predicted labels} \\
        \hline
        1 & [Bình Dương] Chuyên Viên Tư Vấn Tài Chính Khách Hàng Có Sẵn - Kênh Liên Kết Bệnh Viện & - Làm online (trong mùa dịch) hoặc tại quầy giao dịch đặt tại các Bệnh Viện Phụ Sản hoặc Khoa sản khi hết dịch- Không đi thị trường , độc quyền tiếp cận, khai thác nguồn khách hàng tiềm năng và có nhu cầu của bệnh viện và từ các đối tác của bệnh viện- Đại diện Công ty TNHH Manulife Việt Nam để giới thiệu, tư vấn về các gói sản phẩm bảo hiểm của Manulife cho khách hàng.- Thu thập thông tin khách hàng, hướng dẫn khách hàng chuẩn bị và nộp hồ sơ bảo hiểm của khách hàng về bộ phận thẩm định, Chăm sóc, hỗ trợ khách hàng- Thời gian làm việc: Thứ 2 - Thứ 7, 8 tiếng/ 1 ngày. Linh động thời gian làm việc theo ca theo trao đổi với Quản lý. (\textbf{English:} - Work online (during the epidemic season) or at transaction counters located at Obstetrics and Gynecology Hospitals when the epidemic is over - Do not go market, exclusively accessing and exploiting potential and demanding customers of the hospital and from the hospital's partners - Representative of Manulife Vietnam Co., Ltd. to introduce and advise on product packages Manulife's insurance for customers.- Collect customer information, guide customers to prepare and submit insurance documents of customers to the appraisal, care and customer support department- Working time: Monday - Saturday, 8 hours/day. Flexible working time according to the exchange with the Manager.) & Bảo hiểm (Insurance), Dịch vụ khách hàng (Customer Service), Tư vấn (Consulting) & Tư vấn (Consulting)\\ 
        \hline
        2 & Sales Engineer - Kỹ Sư Kinh Doanh (Làm Việc Tại HCM Và Hà Nội) & Phụ trách kinh doanh Ngành hàng Máy, Công Cụ dụng cụ và hóa chất. Phụ trách phát triển quan hệ khách hàng. Lên kế hoạch bán hàng, thực hiện kế hoạch nhằm đạt mục tiêu doanh số hàng tháng. Theo dõi, đối chiếu công nợ với khách hàng, thu công nợ và giao hàng đúng hạn. (\textbf{English:} In charge of sales of Machines, Tools and Chemicals. Responsible for customer relationship development. Make sales plan, execute the plan to achieve the monthly sales target. Follow up, compare debts with customers, collect debts and deliver on time.) & Bán hàng / Kinh doanh (Sales / Business Development), Cơ khí / Ô tô / Tự động hóa (Mechanical / Auto / Automotive), Dầu khí (Oil / Gas) & Tài chính / Đầu tư (Finance / Investment)\\ 
        \hline
        3 & Nhân Viên Kinh Doanh Golf & Tư vấn cho khách hàng về sản phẩm dịch vụ vủa công ty theo data ( Các sản phẩm thiết bị chơi Golf, dự án sân Golf, phòng Golf 3D) Chăm sóc khách hàng Lập báo giá gửi khách hàng, chốt đơn cho khách Tiếp khách hàng mua trực tiếp tại công ty (\textbf{English:} Advising customers on the company's products and services based on data (Golf equipment products, Golf course projects, 3D Golf rooms) Customer care Making quotes for customers, closing orders for customers Receiving guests goods purchased directly at the company) & Tư vấn (Consulting) & [] \\
        \hline
    \end{tabular}
\end{table*}

\section{Conclusion}
\label{conclusion}
In this paper, we propose a methodology for predicting job titles based on the job description texts by constructing a dataset containing nearly 22,000 job descriptions, and implement the Bi-GRU-LSTM-CNN models for the task of job prediction, which is categorized as multi-label text classification task. We use different of pre-trained embedding, and the transformer pre-trained models give better results than other approaches. The BERT with multilingual pre-trained model obtained the highest result on the dataset, which are 62.20\% on the development set, and 47.44\% on the test set. 

From the error analysis, we found that the job descriptions are not enough for predicting relevant job titles in this task. To boost up the accuracy of classification model, the model need to reference to extra information such as job name, job level, and job requirements. Therefore, our future works are research on joint models and multi-task learning methodology for the job title prediction task to improve the accuracy of the classification models. Moreover, we collect more job descriptions about minority job titles to make the dataset more diverse as well as refine the dataset to improve the reliability.

\bibliographystyle{IEEEtran}
\bibliography{references}

\appendix
\label{appendix_job_title}
List of job titles in the dataset: \\
1. An Ninh / Bảo Vệ (Security) \\
2. An toàn lao động (HSE) \\
3. Biên phiên dịch (Interpreter/ Translator) \\
4. Bán hàng / Kinh doanh (Sales / Business Development) \\
5. Bán lẻ / Bán sỉ (Retail / Wholesale) \\
6. Bưu chính viễn thông (Telecommunications) \\
7. Bảo hiểm (Insurance) \\
8. Bảo trì / Sửa chữa (Maintenance) \\
9. Bất động sản (Real Estate) \\
10. CNTT - Phần cứng / Mạng (IT - Hardware / Network) \\
11. CNTT - Phần mềm (IT - Software) \\
12. Chăn nuôi / Thú y (Animal Husbandry / Veterinary) \\
13. Chứng khoán (Securities) \\
14. Công nghệ sinh học (Biotechnology) \\
15. Công nghệ thực phẩm / Dinh dưỡng (Food Tech / Nutritionist) \\
16. Cơ khí / Ô tô / Tự động hóa (Mechanical / Auto / Automotive) \\
17. Du lịch (Tourism) \\
18. Dược phẩm (Pharmacy) \\
19. Dầu khí (Oil / Gas) \\
20. Dệt may / Da giày / Thời trang (Textiles / Garments / Fashion) \\
21. Dịch vụ khách hàng (Customer Service) \\
22. Giáo dục / Đào tạo (Education /Training) \\
23. Giải trí (Entertainment) \\
24. Hàng gia dụng / Chăm sóc cá nhân (Household / Personal Care) \\
25. Hàng hải (Marine) \\
26. Hàng không (Aviation) \\
27. Hành chính / Thư ký (Administrative / Clerical) \\
28. Hóa học (Chemical Eng.) \\
29. In ấn / Xuất bản (Printing / Publishing) \\
30. Khoáng sản (Mineral) \\
31. Kiến trúc (Architect) \\
32. Kế toán / Kiểm toán (Accounting / Auditing / Tax) \\
33. Lao động phổ thông (Unskilled Workers) \\
34. Luật / Pháp lý (Law / Legal Services) \\
35. Lâm Nghiệp (Forestry) \\
36. Môi trường (Environmental) \\
37. Mới tốt nghiệp / Thực tập (Entry Level / Internship) \\
38. Mỹ thuật / Nghệ thuật / Thiết kế (Arts / Creative Design) \\
39. Ngành khác (Others) \\
40. Ngân hàng (Banking) \\
41. Nhà hàng / Khách sạn (Restaurant / Hotel) \\
42. Nhân sự (Human Resources) \\
43. Nông nghiệp (Agriculture) \\
44. Nội ngoại thất (Interior / Exterior) \\
45. Phi chính phủ / Phi lợi nhuận (NGO / Non-Profit) \\
46. Quản lý chất lượng (Quality Control - QA/QC) \\
47. Quản lý điều hành (Executive management) \\
48. Quảng cáo / Đối ngoại / Truyền Thông (Advertising / PR / Communications) \\
49. Sản xuất / Vận hành sản xuất (Manufacturing / Process) \\
50. Thu mua / Vật tư (Purchasing / Merchandising) \\
51. Thư viện (Library) \\
52. Thống kê (Statistics) \\
53. Thủy lợi (Irrigation) \\
54. Thủy sản / Hải sản (Fishery) \\
55. Thực phẩm \& Đồ uống (Food \& Beverage - F\&B) \\
56. Tiếp thị / Marketing (Marketing) \\
57. Tiếp thị trực tuyến (Online Marketing) \\
58. Truyền hình / Báo chí / Biên tập (TV / Newspaper / Editors) \\
59. Trắc địa / Địa Chất (Surveying / Geology) \\
60. Tài chính / Đầu tư (Finance / Investment) \\
61. Tư vấn (Consulting) \\
62. Tổ chức sự kiện (Event) \\
63. Vận chuyển / Giao nhận / Kho vận (Freight / Logistics / Warehouse) \\
64. Xuất nhập khẩu (Import / Export) \\
65. Xây dựng (Civil / Construction) \\
66. Y tế / Chăm sóc sức khỏe (Medical / Healthcare) \\
67. Điện / Điện tử / Điện lạnh (Electrical / Electronics) \\
68. Đồ gỗ (Wood) \\

\end{document}